\documentclass[letterpaper, 10 pt, conference]{ieeeconf}  

\IEEEoverridecommandlockouts                              

\usepackage{graphics} 
\usepackage{graphicx}
\usepackage{times} 
\usepackage{amsmath} 
\usepackage{amssymb}  
\usepackage{xspace}
\usepackage{tabularx}
\usepackage[normalem]{ulem}
\usepackage{xcolor}
\usepackage[normalem]{ulem}

\usepackage{flushend}

\title{\LARGE \bf Local Path Planning among Pushable Objects\\based on Reinforcement Learning}

\author{Linghong Yao, Valerio Modugno, Andromachi Maria Delfaki,\\ Yuanchang Liu, Danail Stoyanov, and Dimitrios Kanoulas
\thanks{$^{1}$The authors are with the Department of Computer Science and Mechanical Engineering, University College London, Gower Street, WC1E 6BT, London, UK. {\tt\small d.kanoulas@ucl.ac.uk}}
\thanks{This work was supported by the UKRI FLF [MR/V025333/1] (RoboHike) and EPSRC [EP/P012841/1].  For the purpose of Open Access, the author has applied a CC BY public copyright license to any Author Accepted Manuscript version arising from this submission.}}

\begin{document}

\maketitle
\thispagestyle{empty}
\pagestyle{empty}

\begin{abstract}
In this paper, we introduce a method to deal with the problem of robot local path planning among pushable objects -- an open problem in robotics. In particular, we achieve that by training multiple agents simultaneously in a physics-based simulation environment, utilizing an Advantage Actor-Critic algorithm coupled with a deep neural network. The developed online policy enables these agents to push obstacles in ways that are not limited to axial alignments, adapt to unforeseen changes in obstacle dynamics instantaneously, and effectively tackle local path planning in confined areas. We tested the method in various simulated environments to prove the adaptation effectiveness to various unseen scenarios in unfamiliar settings. Moreover, we have successfully applied this policy on an actual quadruped robot, confirming its capability to handle the unpredictability and noise associated with real-world sensors and the inherent uncertainties present in unexplored object pushing tasks.
\end{abstract}

\section{INTRODUCTION} \label{sec:intro}
Mobile robots have gained great capabilities in the past decade, such that they are now able to autonomously navigate efficiently and safely even in clutter environments, by avoiding static or dynamically moving obstacles~\cite{Cai2020}. Although, the navigation problem where objects can be moved around to free space --also known as Navigation Among Movable Obstacle (NAMO)-- is still an open problem. This concept mirrors the human instinct to shift, for instance, furniture or other objects blocking their way in densely furnished areas, suggesting that robots could similarly optimize their routes by strategically moving obstacles to clear a path towards their destination. The applications in robotics are highly relevant in several scenarios, including tasks with regular maintenance in factories where unused containers and boxes might block access points, domestic service robots navigating through furniture-cluttered homes, or robots conducting inspections in subterranean environments obstructed by rocks and debris. The capacity for effective obstacle manipulation can greatly enhance autonomous navigation efficiency in these settings.

The problem when movable objects need to be moved around, even in simplified versions, is proved to be NP-hard~\cite{demaine2000pushpush}. To solve global path planning among movable obstacles, previous studies have explored iterative and recursive algorithms~\cite{stilman2008planning}, but often relying on certain simplifications, such as having prior knowledge about the environment, planning tasks offline which is exponential to the number of local obstacles~\cite{moghaddam2016planning}, and limiting movements to axial-aligned object pushes. On the other hand, local path planning in a movable object setting has been minimally studied in the past, with only a few considering sensor inaccuracies when dealing with unexpected object dynamics~\cite{kakiuchi2010working, scholz2016navigation, Armleder2024}, usually based on traditional optimization methods, that require intensive fine tuning and handcrafted design choices.

\begin{figure}[t!]
    \centering
    \includegraphics[width=\linewidth]{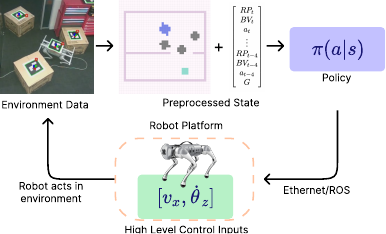}
    \caption{Visual sensors capture the surroundings, and this data is then processed into a specific state representation. This processed information is inputted into a previously trained policy network. The network then generates a strategic action directive for the robot, enabling it to navigate and address the task of local path planning.}
    \label{Fig:workflow}
\end{figure}

\begin{figure*}[t!]
    \centering
    \includegraphics[width=0.9\textwidth]{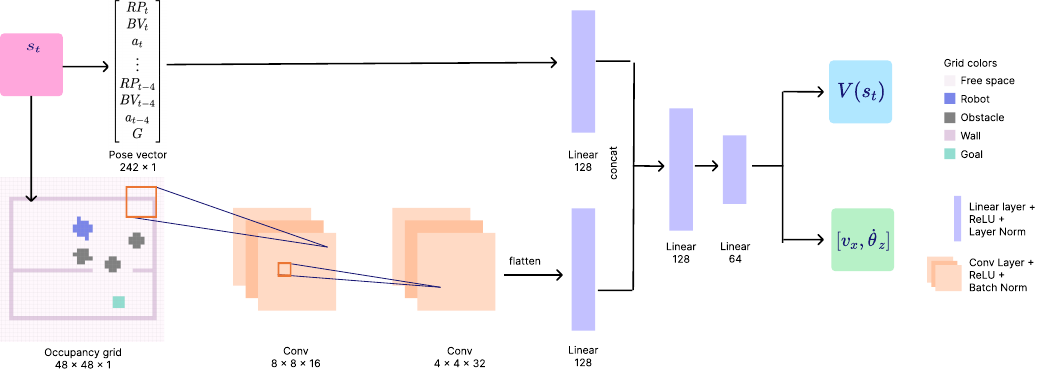}
    \caption{Our approach utilizes a deep neural network for policy-making. The state representation, $s_t$, includes both a vector and a grid. The vector, of length $242$, contains data on the agent's current position, the corners of the object, the previous action taken, and the destination. The grid is a $48\times48$ matrix, with each cell semantically annotated. Initially, the vector is processed through a linear unit, while the grid undergoes processing through two convolutional units followed by a linear unit. The results from both processes are then merged and further processed through two additional linear units. Subsequently, the network splits, employing two distinct sets of weights to generate both the estimated value and the proposed action.}
    \label{Fig:network}
\end{figure*}

In this paper, we propose a novel approach to overcome the aforementioned limitations, by employing deep Reinforcement Learning (DRL) as depicted in Fig.~\ref{Fig:workflow}. In particular, we utilize a neural network for policy-based RL, to allow an online agent complete local path plans without the constraint of previous work for axis-aligned pushing, as well as allowing uncertainties in sensor inputs and obstacle dynamics. Our solution targets the keyhole problem~\cite{stilman2005navigation}, where a mobile robot aims to traverse from one disjointed area to another by pushing obstacles through narrow passages. We acknowledge that other forms of object manipulation exist, such as pulling or lifting, but additional robotic mechanisms such as arms are needed, while just pushing remains NP-Complete. The goal of the trained policy is local path planning that can be integrated into other global navigation methods, such as A*~\cite{hart1968formal}. Our approach is based on Advantage Actor-Critic~\cite{mnih2016asynchronous}, leveraging the advanced capabilities of the NVIDIA Isaac Gym physics engine~\cite{makoviychuk2021isaac} for simulating and training parallel agents.  We showcase outcomes for policies that are adept at navigating through both familiar environments with new movable obstacle placements, and entirely unknown environments with unseen movable obstacles. Moreover, we validate our findings through practical experiments using a Unitree Go1 quadruped robot, illustrating the policy's effectiveness in dealing with sensor inaccuracies and varying dynamics of real-world obstacles.

In Sec.~\ref{sec:rw}, we discuss the literature on the NAMO problem, while in Sec.~\ref{sec:method} we state the problem formulation in the reinforcement setting, and how we implement training in simulation. In Sec.~\ref{sec:exp} we present our results both in simulation and with real robot experiments. Lastly, Sec.~\ref{sec:fw} concludes the paper and points to future directions for solving NAMO with RL.

\section{RELATED WORK} \label{sec:rw}
Path planning, both globally or locally, for obstacle avoidance has been heavily studied in the past~\cite{Liu2023}, and while several interesting techniques use reinforcement learning~\cite{Shen2023} we will not extend further in this related work. In contrary, we will briefly review methods that deal with movable obstacles, especially in the local path planning case.

Global navigation among movable objects/obstacles is a topic that has been studied from the 1980's~\cite{wilfong1988motion} -- an NP-Complete problem even in the simple case of moving square blocks in the plane~\cite{demaine2000pushpush}. A series of papers by Stilman et al., such as~\cite{stilman2005navigation, stilman2008planning}, considered the problem as a graph planning one in which disjointed free spaces (nodes in the graph) can be connected when obstacles can be moved around. In other works, similar setups were solved using RRTs and adaptive heuristics~\cite{nieuwenhuisen2008effective} or axis-aligned obstacle movements~\cite{berg2009path, Raghavan2021}. The problem has been studied in more generic ways, e.g., axis-aligned object manipulations via non-linear optimization~\cite{moghaddam2016planning}. In such solutions, all computation is offline (exponential to the number of objects) with prior knowledge of the environment, including the movability of the objects themselves. In contrary, there are also methods that could online re-plan pushing actions in unknown environments~\cite{wu2010navigation} or by pick-and-place on humanoids~\cite{kakiuchi2010working} using traditional path planning techniques. Hierarchical RL was used first by Levihn et al.~\cite{levihn2013hierarchical} to deal with uncertain sensory information, while later in non-axial manipulation of obstacles developed in~\cite{scholz2016navigation}, a physics-based RL framework in unexpected obstacle behaviors such as rotation was handled. Compared to those, our method runs in constant time complexity and therefore completes similar tasks five times faster, while we are able to solve harder non-linear problems too.  More recently, tactile sensing is used for negotiation of unknown objects~\cite{Armleder2024}, while curriculum learning is used in~\cite{Wang2023} to sole the global navigation among movable obstacles problem. Further extensions to NAMO, such as socially aware obstacle placement, have also been examined in~\cite{renault2019towards, Ellis2022IROS, Ellis2022Access}, using classic search-based approaches.

In this paper, we utilize deep RL to deal with the local path planning problem in narrow spaces. In a similar setup, Xia et al.~\cite{xia2020interactive} used deep RL to deal with collisions with pushable objects, rather than actual planning interactions with those. Given all the aforementioned methods, the real novelty of our approach is:
\begin{itemize} 
  \item We propose a deep RL policy that can solve local path planning among pushable objects, with non-axial-aligned pushing and constant computational complexity.
  \item We demonstrate that the proposed policy is able to work for unseen object positions in known environments, and generalizes to unseen object positions in unknown environments.
  \item We show that reliable sim-to-real transfer is possible to handle sensor noises and uncertain object dynamics.
\end{itemize}

\section{METHODS} \label{sec:method}
In a standard episodic Reinforcement Learning (RL) scenario, an agent interacts with its environment in discrete steps. At each step, the agent observes a state $s_t$ and chooses an action $a_t$ based on a policy distribution $\pi(a_t|s_t;\boldsymbol{\theta})$, with $\boldsymbol{\theta}$ representing the parameters of a function approximator. Following the action, the agent is presented with a new state $s_{t+1}$ and a scalar reward $r_t$. The goal in a policy-based framework is to adjust $\boldsymbol{\theta}$ to enhance the expected total reward $R_t = \sum^{\infty}_{k=0}\gamma^{k}r_{t+k}$, where $\gamma \in (0, 1]$ serves as the discount factor.  Within Advantage Actor-Critic techniques, the approach involves calculating both a policy $\pi(a_t|s_t;\boldsymbol{\theta})$ that dictates action selection and a value function $V(s_t;\boldsymbol{w})$ that predicts the expected reward $\mathbb{E}_{\pi}[R_t|s_t=s]$ if the agent follows policy $\pi$ from state $s_t$. The value function, or critic, is adjusted through the parameters $\boldsymbol{w}$, while the policy, or actor, is modified using the parameters $\boldsymbol{\theta}$.

We developed a deep neural network to serve as a function approximator, modifying its parameters through the process of stochastic gradient descent. The update rules for the parameters are as follows:
\begin{equation}
    \boldsymbol{\theta} \leftarrow \boldsymbol{\theta} + \alpha \nabla_{\boldsymbol{\theta}}\log{\pi(a_t|s_t;\boldsymbol{\theta}})A(s_t, a_t; \boldsymbol{w})
    \label{policy_update}
\end{equation}
\begin{equation}
    \boldsymbol{w} \leftarrow \boldsymbol{w} + \alpha_w \nabla_{w}V(s_t;\boldsymbol{w})A(s_t, a_t; \boldsymbol{w})
    \label{value_update}
\end{equation}
Here, $\alpha$ and $\alpha_w$ denote the learning rates for the policy and value function updates, respectively. The term $A(s_t, a_t; \boldsymbol{w})$ calculates the n-step advantage for a given state-action pair ($s_t$, $a_t$), incorporating a lookahead of $k$ steps, where $A(s_t, a_t; \boldsymbol{w}) = \sum^{k-1}_{i=0}\gamma^{i}r_{t+i}+\gamma^{k}V(s_{t+k};\boldsymbol{w})-V(s_t;\boldsymbol{w})$.

The architecture combines the actor and critic parameters $\boldsymbol{\theta}$ and $\boldsymbol{w}$ through a shared weights strategy as depicted in Fig.~\ref{Fig:network}, enhancing the stability of the learning process. Furthermore, we integrate an entropy term $\nabla_{\theta}H(\pi(s_t;\boldsymbol{\theta}))$ into the update equations Eqs.~\eqref{policy_update} and~\eqref{value_update}, following the guidance of prior research to regularize the learning phase. Additionally, we implement a clipped surrogate objective as advised in recent studies~\cite{mnih2016asynchronous}:
\begin{equation*}
    L^{CLIP}(\theta) = \hat{\mathbb{E}}_t\left[ \text{min}(r_t(\theta)\hat{A}_t, \text{clip}(r_t(\theta), 1-\epsilon, 1+\epsilon)\hat{A}_t)\right]
\end{equation*}
where $\epsilon$ represents a hyper-parameter that needs adjustment. Such methodologies further refine the stability and efficiency of policy updates during training.

\subsection{Problem Formulation}
We address the problem of local path planning, aiming to link two disjointed, neighboring free spaces divided by obstacles that can be pushed around. An agent is introduced to operate within this environment across multiple timesteps, taking actions in a continuous space defined by forward velocity $v_x$ and angular velocity $\dot{\theta}z$. At every timestep, the agent earns a reward $r_t$ from the environment, with the goal being to optimize the total return $\sum^{\infty}_{k=0}\gamma^{k}r_{t+k}$.

The setting is a confined area featuring a narrow path blocked by various pushable objects; the agent/robot needs to move from one place to another by locally pushing objects around.  The episode concludes when the agent either reaches the objective or surpasses the maximum duration allowed for an episode. We presuppose the availability of certain preliminary inputs for defining the agent's state: a semantically annotated, coarse occupancy grid accessible through LiDAR or RGB-D sensors coupled with semantic segmentation techniques~\cite{rosinol2020kimera}; bounding boxes around obstacles detected via object recognition software~\cite{bundlesdfwen2023}; and data on the agent's instantaneous condition derived from the robot's built-in sensors, acknowledging some degree of inaccuracy in all sensory data. Utilizing these initial inputs, we formulate the state of the agent to encompass the target location $G$, the coordinates of pushable obstacles $BV_t$, details on the robot's present status (position, velocity, rotation, angular velocity) $RP_t$, the most recent action $a_t$, and a semantically annotated occupancy map. To impart temporal context to the agent, we incorporate a sequence of prior states concerning $RP_t$, $BV_t$, and $a_t$, echoing the approach used in previous research~\cite{mnih2015human}. Our choice to integrate the last $5$ frames balances effectively between operational efficiency and computational demands. The comprehensive state consists of the occupancy map and a vector detailing $G$, $RP_{t-4:t}$, $BV_{t-4:t}$, and $a_{t-4:t}$, as illustrated in Fig.~\ref{Fig:network}. For the policy to remain functional and adaptable, we employ a straightforward control scheme based on a unicycle model, which is characterized by two movement parameters: $v_x$ (linear velocity) and $\dot{\theta}_z$ (angular velocity). This policy is designed to be versatile, applicable across various mobile robotic platforms, as demonstrated in the experimental section.

\begin{table}[t!]
\caption{REWARDS AT EACH TIMESTEP}
\centering 
\setlength{\tabcolsep}{4pt}
\begin{tabular}{l l c } 
reward & description & weight\\ 
\hline & \\[-1.5ex]
goal  & 1 if reach goal, 0 otherwise & 10   \\
progress   & [-1, 1] $\propto$ velocity towards goal & 1   \\
dist  & [0, 1] $\propto$ distance to goal & 0.1   \\
wall collision  & -1 if collision with wall & 0.2   \\
object collision  & -1 if collision with object & 0.1   \\
vel effort  & $ [-1, 0] \propto v_{target}$ target velocity & 0.05   \\
rot effort  & $ [-1, 0] \propto \dot{\theta}_{target}$ & 0.1   \\
vel offset & $ [-1, 0] \propto |v_{actual}-v_{target}|$  & 0.2 \\
rot offset & $ [-1, 0] \propto |\dot{\theta}_{actual}-\dot{\theta}_{target}|$  & 0.1 \\
time  & -1 & 1   \\
\end{tabular}
\label{table:rew} 
\end{table}

The goal for the agent is to optimize the total cumulative reward. Rewards at each step are outlined in Table~\ref{table:rew}. A significant reward is allocated for successfully completing the objective, with no reward given for incomplete tasks. Incremental positive rewards are awarded each timestep for actions that advance the agent toward the target (\textit{progress}), or for maintaining proximity to the goal (\textit{dist}). Conversely, minimal negative rewards are assigned in relation to the effort required for movement (\textit{vel effort}, \textit{rot effort}), collision with walls (\textit{wall collision}), and interactions with objects while pushing (\textit{object collision}). Additionally, the agent is penalized for significant deviations between intended and executed actions, often caused by object collisions or sudden changes in action (\textit{vel offset}, \textit{rot offset}).

\subsection{Implementation}
Our agent's training is conducted in a simulated environment using NVIDIA Isaac Gym~\cite{makoviychuk2021isaac}, a platform that supports the concurrent training of multiple agents on a single GPU, promoting both stability and efficiency in policy-based learning strategies~\cite{mnih2016asynchronous}. We adopt the advantage actor-critic method, utilizing a deep neural network that integrates two distinct components of the state space, as depicted in Fig.~\ref{Fig:network}. This state space is comprised of both vector and grid elements, each normalized within the range of $[-1, 1]$. The vector portion undergoes processing in a linear block, which includes a single perceptron layer with $128$ units, followed by ReLU activation and a normalization layer. Meanwhile, the grid data is processed through two convolutional blocks, each equipped with convolutional filters, ReLU activation, and batch normalization. Specifically, the first convolutional block utilizes $16$ filters sized $8\times8$, and the second employs $32$ filters of $4\times4$. After flattening the output, it is directed through a linear block containing $128$ units. Following the concatenation of the two processed streams, the combined data is further processed through two linear blocks, one with $128$ units and another with $64$ units, leading to the final stage where two separate linear layers generate the value estimate $V(s_t)$ and the proposed action $[v_x, \dot{\theta}_z]$.

\begin{figure}[t!]
    \centering
    \includegraphics[width=\linewidth]{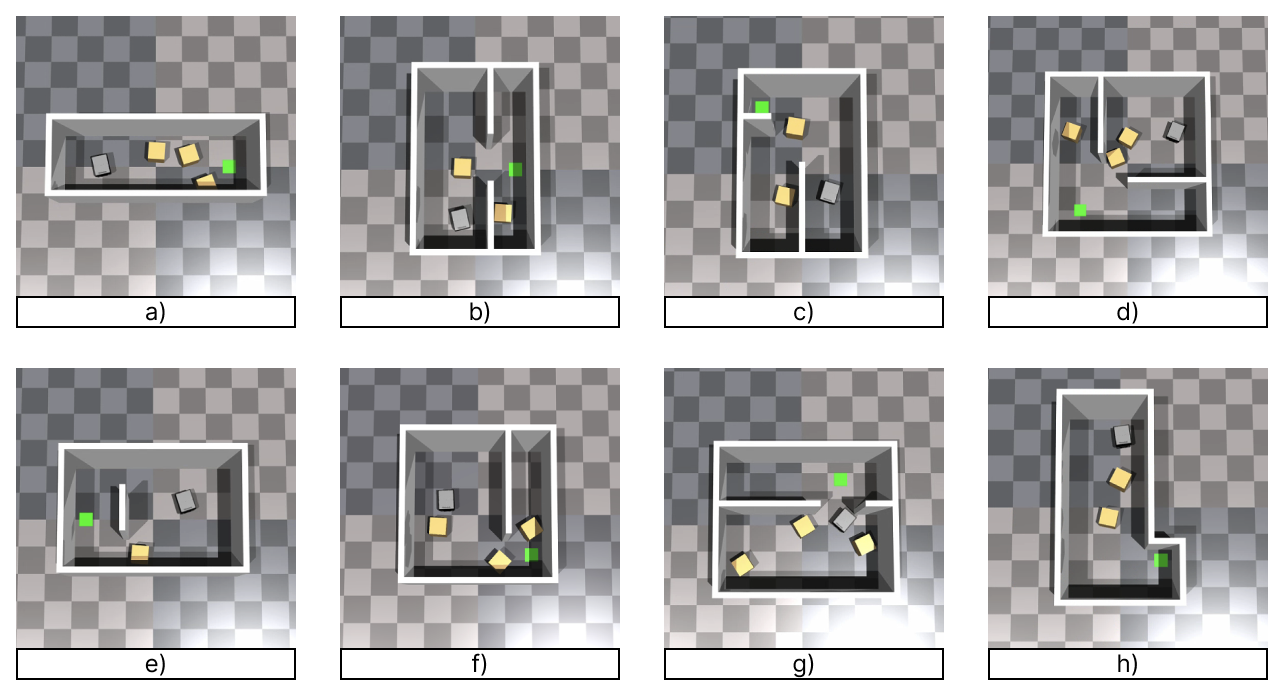}
    \caption{Fixed maps and random obstacle positions. Agents (gray) need to push object (yellow) to reach the goal (green). The maps include corridors (a,b), mid (c,d), side (e,f), and diagonal (g,h) doorways.}
    \label{Fig:rooms}
\end{figure}

\subsection{Scene Generation and Curriculum Training}
In Fig.~\ref{Fig:rooms}, we present eight distinct map layouts created within Isaac Gym, each designed to encapsulate a broad spectrum of local Navigation Among Movable Obstacles path planning among pushable objects scenarios. These configurations include tight corridors (maps $a$ and $b$), entryways with adjacent spaces (maps $c$ and $d$), entrances flanked by walls (maps $e$, $f$, and $i$), and diagonal doorways (maps $g$ and $h$). Agents are placed in specifically designed rooms, each measuring approximately $6\times6$ square meters, a dimension that accommodates mobile robots ranging from $0.5$ to $1$ meter in length, enabling their navigation through narrow paths approximately $1$ to $2$ meters wide. The challenge is heightened by obstacles placed close to these narrow paths, complicating the robot's ability to traverse them. Robots and their target destinations are randomly positioned within designated zones, though there's a slight chance (for instance, $5\%$) that robots may spawn anywhere on the map, ensuring they have the opportunity to explore every area to some extent.

\begin{figure}[t!]
    \centering
    \includegraphics[width=1.0\linewidth]{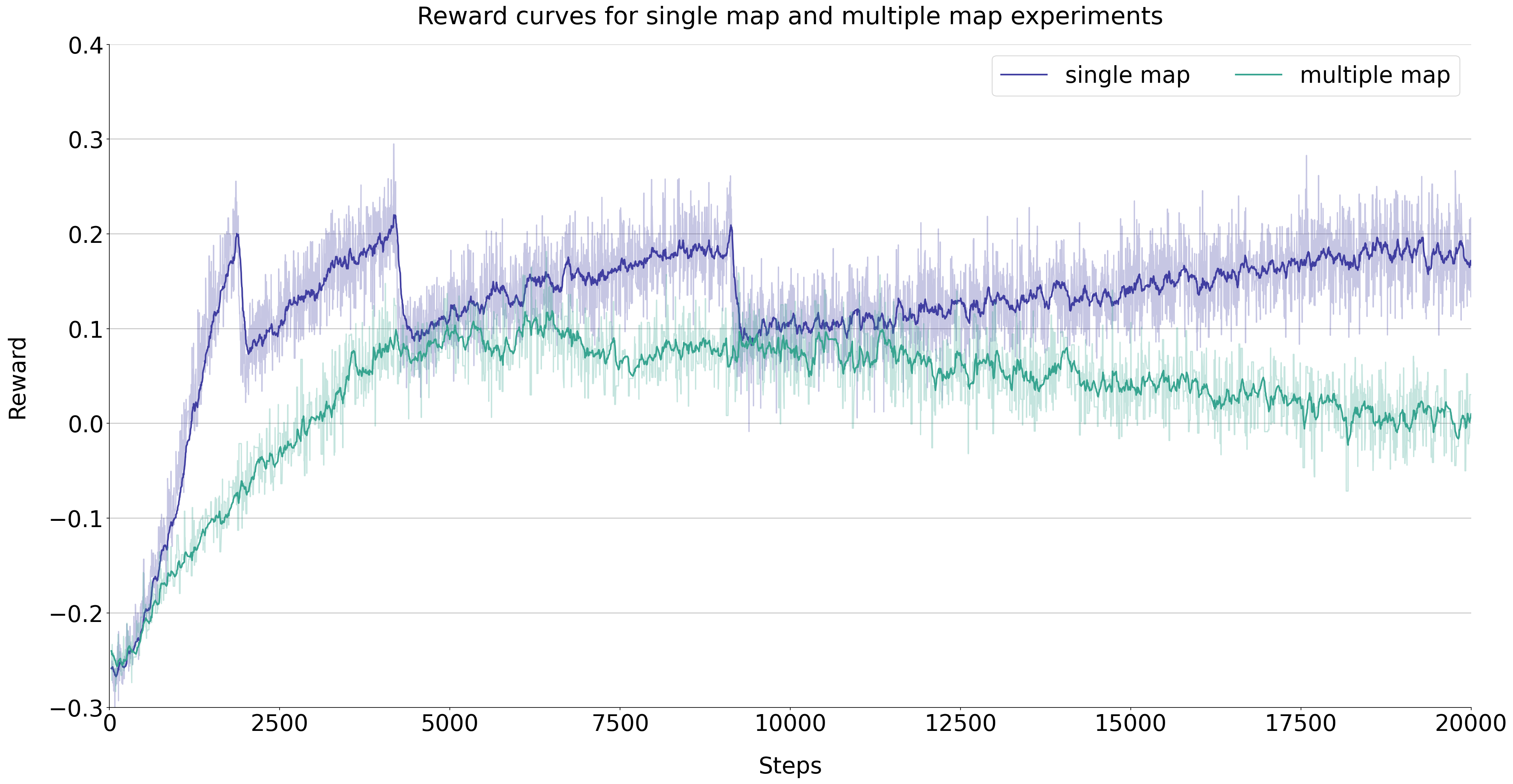}
    \caption{The progression of rewards through policy update iterations is depicted with the single map scenario in purple and the multi-map scenario in green. The phenomenon of curriculum learning is evident through the steep declines in the purple line, indicating the increasing difficulty of tasks; this effect is mitigated in the multi-map scenario, where the curve is more gradual since not all tasks advance through the curriculum at the same time. While the single map approach tends toward a steady state of balance, the multi-map approach demonstrates a need for early termination due to its varied progression.}
    \label{Fig:curve}
\end{figure}

We opt for pushable obstacles (represented as boxes) approximately $60cm^3$ in size, limiting the number to a maximum of $5$ per room. This quantity was selected as it populates the room with sufficient diversity without overcrowding it, which would complicate the generation of object positions. Random generation of obstacle locations would simplify most scenarios, minimizing the need for the agent to interact with obstacles. To counteract this, obstacles are placed with a skewed probability towards more challenging positions. Specifically, for each obstacle $i$, there's a $\lambda p_i$ chance it will appear at a random spot within the room, with an arbitrary orientation. With a $\lambda(1-p_i)$ likelihood, it is positioned in strategically difficult locations (e.g., blocking a narrow path or situated close to such a path) with slight random adjustments. An obstacle is omitted, represented as zeros in the input data, with a $1-\lambda$ probability. The probability $p_i$ for each object ranges from $0.2$ to $0.6$, and $\lambda$ is incrementally increased from $0.2$ in set steps. Through the random placement of pushable objects, we compel the agent to discover solutions for varying obstacle configurations during each attempt. This process encourages the agent to develop an understanding of how different obstacles relate to one another, including determining the sequence in which to move the obstacles.

\begin{figure*}[t!]
    \centering
    \includegraphics[width=0.8\textwidth]{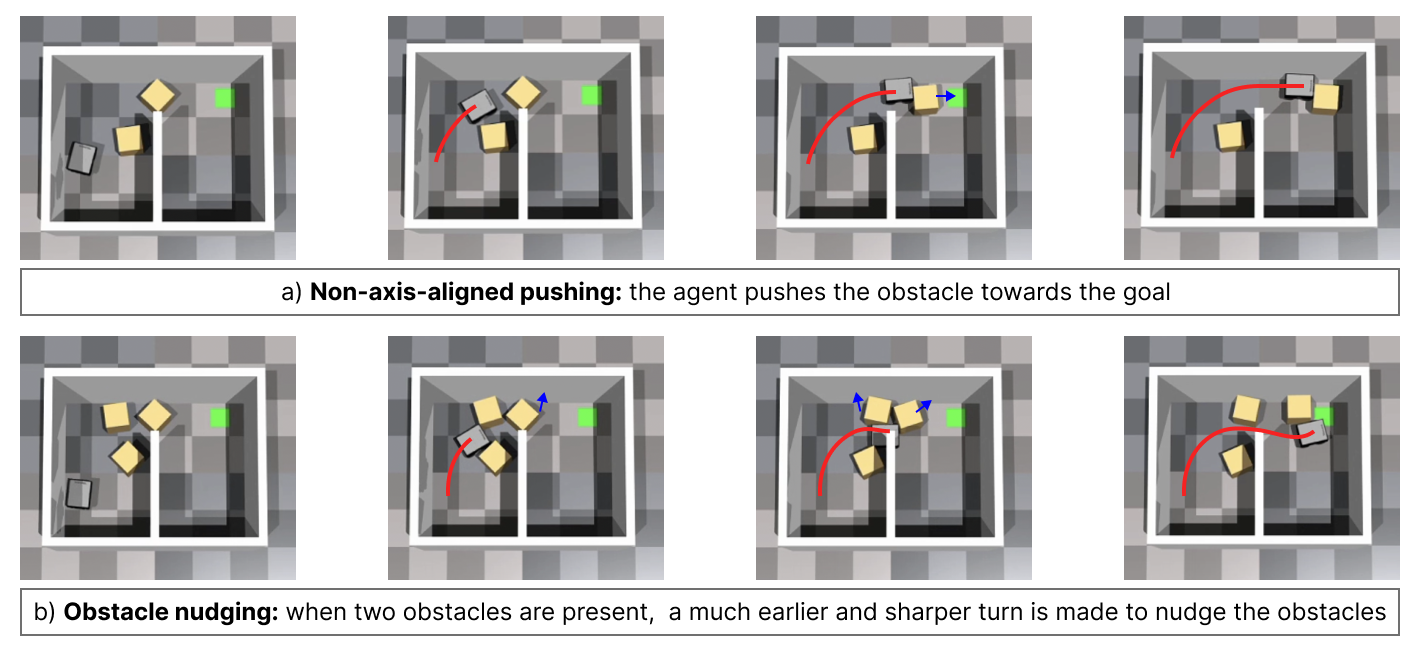}
    \caption{Qualitative simulated results on single map setting: the agent adapts to different obstacle positions and finds efficient paths.}
    \label{Fig:experiment_1}
\end{figure*}

The parameter $\lambda$ serves as a mechanism to modulate the number of objects, thus adjusting the complexity of the Navigation Among Movable Obstacles path planning among pushable objects challenge. This methodology introduces a curriculum training strategy, where the difficulty escalates as the agent becomes proficient at simpler tasks. A completion rate threshold (e.g., $90\%$) determines when the difficulty level increases. Starting from $\lambda=0.2$, the agent is trained until it achieves a $90\%$ completion rate, at which point $\lambda$ is incremented by $0.2$. This cycle repeats until $\lambda$ equals $1$, ensuring that $4$ to $5$ objects are consistently present at the commencement of each scene. The impact of this curriculum learning approach throughout the training is illustrated in Fig.~\ref{Fig:curve}.

\subsection{Domain Randomization}
Intensive domain randomization is employed across both state and action spaces, incorporating Gaussian white noise to reflect the anticipated variability and noise in real-world sensor readings and object movements. This approach is designed to ensure that the developed policy can effectively adapt to and manage the unpredictability inherent in real-world scenarios. By integrating noise into the state space inputs, we aim to mimic the sensor inaccuracies robots might face, such as variations in the detection of obstacle and robot positions over time. It's important to note that noise is added separately to the vector and grid states, mirroring the often uncorrelated nature of real-world sensor noise sources. Likewise, we introduce Gaussian noise to the model's action outputs to replicate unforeseen robot dynamics. The adoption of domain randomization techniques in our training methodology reduces the discrepancy between simulation and real-world applications, thereby enhancing the policy's reliability in the presence of noisy data and unpredictable dynamics. This preparation ensures policy robustness, a claim we will substantiate in the subsequent section.

\section{EXPERIMENTAL VALIDATION} \label{sec:exp}
\subsection{Simulation}
Initially, we evaluate the effectiveness of our trained policy through both numerical and observational assessments in a simulated environment. The training is conducted under two distinct scenarios: a single room and multiple rooms. In the single-room scenario, the agent's training involves navigating through randomly placed objects, as depicted in Fig.~\ref{Fig:experiment_1}. Conversely, the multiple-room scenario introduces the agent to eight varied room configurations. Our findings indicate that within the single-room setup, the policy successfully adapts to novel obstacle arrangements within a familiar setting, showcasing efficient strategies and a minimal rate of failures. Meanwhile, in the context of multiple rooms, the policy demonstrates adaptability to new obstacle placements across unfamiliar settings, albeit with slightly less refined behaviors.

For both experimental setups, we implement identical training configurations. The methodology for scene creation and the introduction of variability are outlined in Sec.~\ref{sec:method}, highlighting the random allocation of agent, target, and pushable object locations, the adoption of a progressively challenging curriculum, and the application of domain randomization to both inputs and outputs. The optimization process leverages the ADAM algorithm~\cite{kingma2014adam}, enhanced with l2 regularization to improve stability~\cite{farebrother2018generalization}, and incorporates gradient clipping to mitigate the risk of exploding gradients, a common issue in scenarios involving off-policy learning, approximate function representation, and value bootstrapping (referred to as the deadly triad)~\cite{sutton2018reinforcement}. An adaptive learning rate strategy is employed, guided by a predefined KL divergence threshold. The training protocol specifies a horizon of $50$ actions, with policy updates occurring every $20$ physics simulation steps (equivalent to $333$ milliseconds). Episodes are capped at a duration of $45$ seconds, translating to $2700$ physics simulation steps. The training process engages $4,000$ parallel environments, with each update drawing on a mini-batch of $2,000$ samples. This regimen is executed on an NVIDIA RTX 3080 GPU, spanning $20,000$ policy update iterations.

\begin{figure*}[t!]
    \centering
    \includegraphics[width=0.7\textwidth]{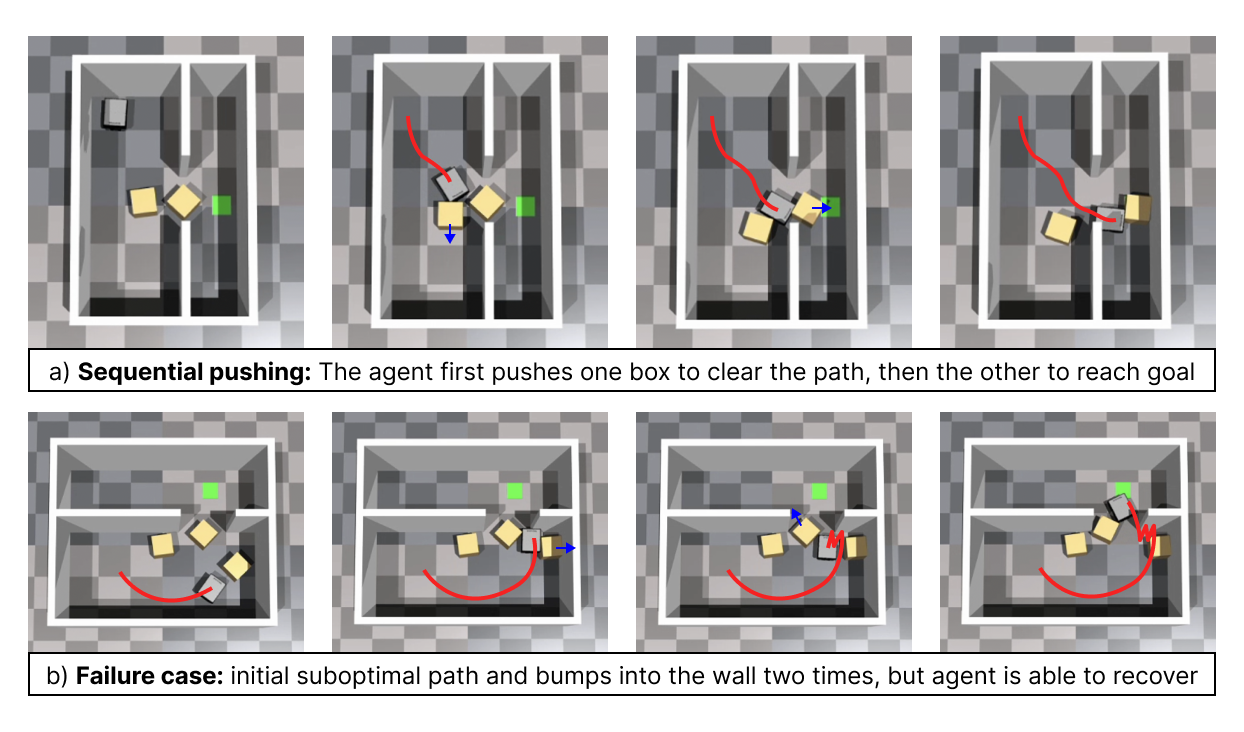}
    \caption{Observational findings from simulations in various map configurations reveal that: a) the agent strategically maneuvers around obstacles in sequence, optimizing its path with only slight deviations; b) the agent effectively adapts to unforeseen changes in dynamics.}
    \label{Fig:experiment_2}
\end{figure*}

\begin{table}[t!]
\caption{POLICY PERFORMANCE ON SINGLE MAP WITH VARYING $\lambda$}
\centering 
\setlength{\tabcolsep}{4pt}
\begin{tabular}{l c c c c } 
$\lambda$ & objects & completion rate & time taken & objects pushed\\ 
\hline & \\[-1.5ex]
0   & 0    & 99.9 & 6.80 & 0  \\
0.2 & 0.7  & 98.5 & 7.13 & 0.34  \\
0.4 & 1.3  & 98.3 & 7.68 & 0.64  \\
0.6 & 2.1  & 97.4 & 8.09 & 1.00  \\
0.8 & 2.8  & 94.5 & 8.14 & 1.37  \\
1   & 3.7  & 91.0 & 8.69 & 1.92  \\
\end{tabular}
\label{table:single} 
\end{table}

In our initial experiment using a singular map layout, the policy undergoes training within a consistent map (Fig.~\ref{Fig:experiment_1}) with varied starting points for the agent, destination, and pushable obstacles. The evaluation is conducted over $1,000$ unique scenarios within the identical map setup but featuring previously unseen obstacle placements. This evaluation process is replicated six times, each instance adjusting the $\lambda$ parameter to increment the obstacle count within the scene, thus escalating the complexity of the challenges faced. Fig.~\ref{Fig:experiment_1} showcases exemplary outcomes from this experiment. Observations from the upper section indicate the agent's capability to maneuver obstacles in a non-linear manner along its path. The lower segment illustrates the agent executing a tighter maneuver near a doorway, opting to slightly displace obstacles to facilitate passage, a strategy necessitated by the potential blockage of the goal (indicated in green) if the obstacles were pushed directly into the doorway. These actions underscore the adaptability and strategic planning of the agent, surpassing the limitations of axis-aligned obstacle manipulation commonly seen in previous research.

The numerical outcomes of this study are detailed in Table~\ref{table:single}, with each row indicating a progressive increase in challenge level corresponding to higher $\lambda$ values. It's important to highlight that the actual count of objects present in each scenario doesn't linearly correlate with $\lambda$, due to the stochastic nature of object placement and the occasional lack of available space for additional objects. As the scenarios grow more complex with an increased presence of objects, we note a decline in the success rate, primarily when the agent encounters situations where pathways or the goal itself become obstructed, rendering progress impossible. In the table's final column, we document the average number of objects displaced during successful task completions. Interestingly, this figure typically represents only half of the average total object count in the scene, suggesting the agent prioritizes moving only those obstacles that directly impede its route. Across the single map experiments, the agent exhibits consistent and effective behavior, adeptly adapting to new obstacle arrangements and achieving a solution in $91\%$ of instances, even under the most demanding conditions.

\begin{table}[t!]
\caption{Comparison between policy performance on training maps and test map, with $\lambda=0.8$}
\centering 
\setlength{\tabcolsep}{4pt}
\begin{tabular}{l c c c } 
Experiment & completion rate & time taken (s) & objects pushed\\ 
\hline & \\[-1.5ex]
Training (8 maps)     & 79.8 & 10.99 & 1.12  \\
Testing  (unseen map)       & 54.3 & 13.05 & 1.51  \\
\end{tabular}
\label{table:comparison} 
\end{table}

\begin{figure*}[t!]
    \centering
    \includegraphics[width=1.0\linewidth]{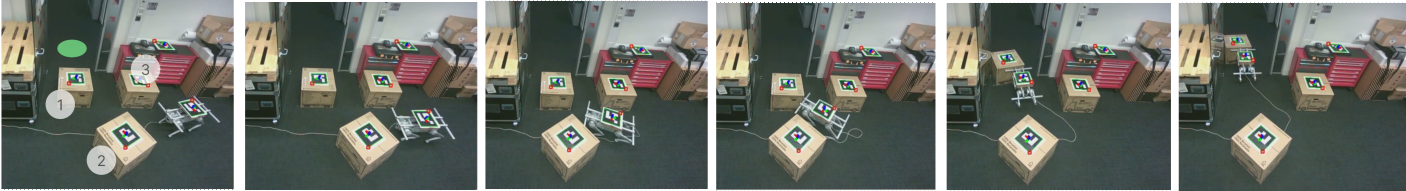}
    \caption{Left to right: \textbf{[t=0]} Green dot is the goal. The robot: \textbf{[t=4]} avoids collision with object 3; \textbf{[t=8]} moves box 2 to create small opening; \textbf{[t=10]} stops pushing object 2 and rotates towards box 1; \textbf{[t=12]} pushes object 1 out of the door. \textbf{[t=23]} Goal reached. The robot performs non-axis-aligned pushing.}
    \label{Fig:experiment}
\end{figure*}

In our subsequent experiment, we assess the ability of a singular policy, trained across multiple map configurations, to adapt not only to novel obstacle placements but also to entirely new environments. The training encompasses maps labeled $a$ through $h$, as depicted in Fig.~\ref{Fig:rooms}, followed by testing the policy in $1,000$ scenarios within an unfamiliar map layout (the same layout utilized in the single map experiment). Maintaining stable learning progress in this multifaceted context proves to be challenging. Despite the implementation of various regularization strategies, achieving consistent policy convergence is not always possible. As illustrated in Fig~\ref{Fig:curve}, the policy's performance fails to match that observed in the single map scenario. Consequently, we resort to early stopping, selecting the most effective policy version identified during the training phase for further application.

Within the context of training across various maps, the agent develops adaptable strategies suitable for diverse scenarios, such as the methodical displacement of obstacles and the ability to dynamically respond to unforeseen events and inaccuracies in sensory input. An instance of this is demonstrated in Fig.~\ref{Fig:experiment_2}-a, where the agent executes non-linear, ordered obstacle movement (preferring to move one object before another). This figure underscores the agent's proficiency in fine-tuning its approach by making only minor deviations from its intended path, a feat difficult to achieve under the limitations of axis-aligned movement restrictions. Fig~\ref{Fig:experiment_2}-b showcases the agent's capability to navigate out of less favorable situations. Challenges in accurately identifying the doorway due to the coarse resolution of input images may lead the agent to mistakenly contact a wall adjacent to a doorway, as depicted in frame~$2$. Nevertheless, the agent effectively maneuvers out of such predicaments by reversing and then proceeding forward again, as illustrated in frames~$3$ and~$4$. Although this approach results in a policy that may not be as refined as that developed under a single map framework, it evidentially highlights the agent's resourcefulness in rectifying its course from disadvantageous positions.

The numerical outcomes are detailed in Table~\ref{table:comparison}, assessing the agent's efficacy within familiar settings (the eight original maps) against novel obstacle configurations and entirely new environments. The agent manages an approximate success rate of $80\%$ across all eight maps in the most challenging scenarios, while it secures a $54\%$ success rate in a novel map setting. Although the performance in the unfamiliar environment is comparatively lower, it nonetheless evidences a notable capability. Enhancing performance further could likely be achieved through increased randomization in training data and leveraging prioritized experience replay to re-engage with more complex path planning problems~\cite{schaul2015prioritized}.

\subsection{Failure Cases}
In scenarios involving a single map, the predominant cause of failure is typically the agent pushing an object to a position from which the path becomes irreversibly obstructed. Conversely, the multi-map approach exhibits a higher incidence of failures, often characterized by the agent making superfluous maneuvers or occasionally becoming ensnared in a corner, as depicted in Fig.~\ref{Fig:experiment_2}. These issues are likely the result of inadequate feature discernment by the neural network, which has been trained on data exhibiting high inter-correlation, coupled with the limited detail available in the input imagery. Enhancements in feature extraction could be achieved by expanding the variety of training environments and employing experience replay techniques to reduce data correlation. Similarly, an increase in the resolution of the input images could further mitigate these issues.

\subsection{Robot Experiments}
This section details our results using an actual quadruped robot (Unitree Go1), outfitted with aluminum extensions to aid in pushing obstacles. We utilize cardboard boxes as pushable objects, each measuring approximately $50\times50\times50cm^3$. Due to the interaction between the boxes and the carpet, the robot is limited to pushing a single box at a time. To monitor the positions of both the obstacles (edges) and the robot (orientation), two overhead cameras equipped with ArUco Markers are employed. This setup allows us to generate a 2D grid that forms the basis of the state space. While the specifics of obstacle recognition and mapping are not covered within this document, we demonstrate the robot's capability to navigate through sensor discrepancies and unpredictable obstacle movements. The decision to apply a neural network trained in a singular map environment is due to its demonstrated reliability and stability, underscoring the feasibility of applying the developed policy in real-world scenarios, characterized by variable robot dynamics and sensor data.

Our evaluations indicate that the robot is adept at maneuvering through confined spaces that include pushable obstacles, as illustrated in Fig.~\ref{Fig:experiment}. The objective for the robot is to locate a designated green point, effectively addressing the Navigation Among Movable Obstacles (local path planning challenge by circumventing collision with box $3$ (by time $t=4$), strategically rotating and slightly moving box $2$ to make way (by time $t=10$), and displacing box $1$ from the entrance until the target is reached (by time $t=23$). It was commonly noted that the robot employs pushing actions along curved paths, utilizing minimal adjustments to alter an obstacle's orientation, frequently resulting in the most efficient method for clearing narrow passages.

\section{CONCLUSIONS} \label{sec:fw}
In our study, we introduced a deep reinforcement learning strategy enabling a robot to execute non-linear obstacle manipulation to address local path planning among pushable objects, maintaining constant computational complexity. This approach was validated both in simulated environments and through practical implementation on a physical robot, accounting for sensor inaccuracies. Looking ahead, our goals include refining the learning process for agents operating across diverse map layouts. Furthermore, we plan to explore strategies enabling agents to navigate around obstacles of uncertain movability and novel geometries by integrating these variables into their training regimen.

\addtolength{\textheight}{-8cm}   


\bibliographystyle{IEEEtran}
\bibliography{iros2024leon}

\end{document}